\documentclass{article}

\usepackage{microtype}
\usepackage{graphicx}

\usepackage{subfigure}
\usepackage{booktabs} 

\usepackage{hyperref}
\usepackage{rotating}
\usepackage{multirow}
\usepackage{amsmath}
\usepackage{amssymb}

\usepackage{hyperref}

\usepackage[accepted]{icml2021}

\begin{document}
	
	
	
	
	%
	%
	%
	%
	\title{Learning Graph Representation by 
		 Aggregating Subgraphs \\via Mutual Information Maximization}
	\author{Chenguang Wang\\
		University of Chinese Academy of Sciences, Beijing, China\\
		{\tt\small wangchenguang19@mails.ucas.ac.cn}
		\and
		Ziwen Liu\\
		University of Chinese Academy of Sciences, Beijing, China\\
		{\tt\small liuziwen18@mails.ucas.ac.cn}
	}
	\date{}
	
	\maketitle
	
	
	
	\begin{abstract}
		In this paper, we introduce a self-supervised learning method to enhance the graph-level representations with the help of a set of subgraphs. For this purpose, we propose a universal framework to generate subgraphs in an auto-regressive way and then using these subgraphs to guide the learning of graph representation by Graph Neural Networks. Under this framework, we can get a comprehensive understanding of the graph structure in a learnable way.
		And to fully capture enough information of original graphs, we design three information aggregators: \textbf{attribute-conv}, \textbf{layer-conv} and \textbf{subgraph-conv} to gather information from different aspects. 
		And to achieve efficient and effective contrastive learning, a Head-Tail contrastive construction is proposed to provide abundant negative samples. Under all proposed components which can be generalized to any Graph Neural Networks, in the unsupervised case, we achieve new state-of-the-art results in several benchmarks. We also evaluate our model on semi-supervised learning tasks and make a fair comparison to state-of-the-art semi-supervised methods.

\end{abstract}

\section{Introduction}
\label{intro}
Graph Neural Networks (GNN) has shown the extensive ability to mine the intrinsic information of graph structure data and been applied to many areas such as social networks~\cite{kipf2016semi}, human activities~\cite{li2019actional}, knowledge graphs~\cite{vivona2019relational} and many more. They aim to learn reasonable low-dimensional representations for nodes and graphs by preserving both network topology structure and node content information as much as possible. Based on the learned representations, we can properly solve various tasks in a deep-learning way such as classification tasks in node-level~\cite{kipf2016semi,hamilton2017inductive,defferrard2016convolutional} and graph-level~\cite{zhang2018end, ying2018hierarchical}. Many researchers solve graph tasks by supervised learning in an end-to-end way~\cite{zhang2018end, ying2018hierarchical, pan2015joint, li2020graph}, however, in many fields, labeling graphs procedurally using strong prior knowledge is costly. As a consequence, unsupervised learning and semi-supervised learning methods are important technics to alleviate this dilemma. One way focuses on
reconstructing the graph structure~\cite{kipf2016variational}, the other is contrastive methods~\cite{hjelm2018learning,velickovic2019deep,sun2019infograph} which make constraints via mutual information.

Most recent approaches aim to maximize the mutual information between graph representation and node representations, to make the learned graph representation more informative and meaningful. It achieved state of the art results in graph-level and node-level downstream tasks~\cite{sun2019infograph,hassani2020contrastive,li2020graph}. InfoGraph~\cite{sun2019infograph} focuses on unsupervised learning and semi-supervised learning which learns by maximizing the mutual information between the concatenated multi-hop node representations and graph representations after a readout function. What's more,~\cite{hassani2020contrastive} uses Graph Diffusion Networks (GDN)~\cite{klicpera2019diffusion} to generate multi-views of the original graph, constructs contrastive samples by sampling subgraphs, and finally maximizes the mutual information between different pairs of graph and nodes.~\cite{li2020graph} deals with graph classification and node classification in the view of multiscale graph neural networks with graph pooling. They use intermediate fusion across scales, and a mutual information-based pooling method called VIPool so that information communication is available among different scales and graphs can be pooled reasonably.

In this paper, we concentrate on how to aggregate the information from the original graphs and how to give a reasonable and better constraint to retain the essential information of graphs. 

To aggregate the information, we separate the aggregation into three aspects: Aggregation for node information (Node-Agg), Aggregation for multi-scales graph information (Layer-Agg), and Aggregation for subgraph information (Subgraph-Agg). From the perspective of Node-Agg, we aim to aggregate the original information from graphs by \textbf{attribute-conv}; from the perspective of Layer-Agg, we aggregate the multi-hop node representations from different GNN layers by \textbf{layer-conv} and from the perspective of Subgraph-Agg, we use \textbf{subgraph-conv} to aggregate the information from the generated subgraphs so that we can get the reconstructed graph representations in a learnable 
way. As for the generation of subgraphs, we propose a universal framework that can generate subgraphs autoregressively in a learnable way. 

To give a reasonable constraint on representation learning, we can guide our model by maximizing the mutual information between the original graph and the reconstructed graph representations rather than between graph and node representations as in previous works. Furthermore, we propose a so-called Head-Tail contrastive method to generate more negative samples so that making stronger constraints on learned representations.
Based on our method, we get state-of-the-art performance on graph classification tasks. Overall, our contributions contain the following components:

 
 \begin{itemize}
	\item A meaningful information aggregation method by simple learnable convolution kernels: \textbf{attribute-conv} used for fusing different kinds of original information; \textbf{layer-conv} used for aggregating different scales of node representations obtained from GNN and \textbf{subgraph-conv} used for getting the reconstructed graph representations;
	\item A universal and learnable framework to generate subgraphs autoregressively. Then by these subgraphs and the \textbf{subgraph-conv}, we can get a reconstructed graph which leads to a reasonable objective function where we maximize the mutual information between the original graph and the reconstructed graph;
	\item A comprehensive construction method of negative samples which we call Head-Tail contrastive sampling used for contrastive learning which can provide a meaningful constraint on the learned representations;
	\item All above components can be seen as plug-and-play modules so that can be transferred to any graph models flexibly. And benefit from these components, we achieve state-of-the-art results which improve performance by a big margin compared to previous works on MUTAG, PTC-MR, REDDIT-BINARY, and REDDIT-MULTI-5K. 
\end{itemize}
\section{Related Work}\label{related work}

\subsection{Unsupervised Representation Learning}\label{uns_sec}
\textbf{Graph kernel methods}~\cite{prvzulj2007biological,kashima2003marginalized,borgwardt2005shortest, shervashidze2009efficient} are a kind of technic commonly used for node classification, which decomposes the graph into several subgraphs and then measures similarities between them. Much work has focused on deciding the most suitable sub-structures by hand-craft similarity measures between sub-structures.

\textbf{Contrastive learning} as a recently popular unsupervised method, has been widely applied in Nature Language Processing (NLP)~\cite{oord2018representation}, computer vision~\cite{chen2020simple,he2020momentum,tian2019contrastive}, of course in Graph Representation learning~\cite{qiu2020gcc,you2020graph,hassani2020contrastive}. The main idea of contrastive learning is to make similar samples closer than different ones, therefore allowing the representations can better correspond to the inputs. And this approach currently performs best in unsupervised graph and node classification. Deep graph infomax (DGI)~\cite{velickovic2019deep}, based on Deep InfoMax (DIM)~\cite{belghazi2018mine,hjelm2018learning}, learns node representations through contrasting node and graph code. Then InfoGraph~\cite{sun2019infograph} additionally combines DGI with Graph Isomorphism Network (GIN)~\cite{xu2018powerful} to learn node and graph representations from different scales. Furthermore, InfoGraph extends to semi-supervised learning and gets outstanding performance. In~\cite{hassani2020contrastive}, they learn representations by contrasting different structural views of graphs and show the state-of-the-art performance in node and graph classification.

\subsection{Semi-supervised Learning}
Semi-supervised learning is a kind of method used in training datasets that is a mixture of labeled and unlabeled data. There are many different approaches to the semi-supervised learning problem, for example, Pseudo-Label method~\cite{lee2013pseudo} regards the prediction of unlabeled data as a pseudo-label of unlabeled data, then trains the network with all data together, and uses a low weight of the loss of the unlabeled data part. Laddar Network~\cite{rasmus2015semi} also gives a strong idea. It combines supervised learning with unsupervised learning in deep neural networks to apply unlabeled data information to supervised learning reasonably.
\subsection{Learning by Mutual Information}
Mutual information is a metric used to measure the correlation between two random variables, which on the other hand expresses the amount of information shared between them. InfoMax~\cite{linsker1988self} aims to learn a representation more informative about input. But computing mutual information is always a notorious problem. Following that, much work has been done to maximize mutual information by kernel method or optimize the lower bounds, for example, Contrastive Predictable Coding (CPC)~\cite{oord2018representation} optimizes it through InfoNce lower bound of mutual information. And then Mutual Information Neural Estimator (MINE)~\cite{belghazi2018mine} and DIM propose a general mutual information neural estimator and have been proven a great performance in representation learning. Due to its interpretability and good performance, DIM has been applied to various fields.

\section{Method}
In this section, we first unify the notations and concepts for the convenience of description. Then we
separate our proposed method into three stages: Aggregation for node information (Node-Agg), Aggregation for multi-scales graph information (Layer-Agg), and Aggregation for subgraph information (Subgraph-Agg). Especially in the Subgraph-Agg stage, for the purpose of obtaining meaningful graph representations from the view of the subgraph, we introduce an auto-regressive method which is a universal self-supervised framework for the graph generation. 

\subsection{Preliminary}
\textbf{Unsupervised Learning on Graphs.} In unsupervised case, given a set of graphs $\mathbf{G}=\{\mathcal{G}^1,\mathcal{G}^2,\dots\}$ without labels ($|\mathbf{G}|$ is the number of graphs batch), we aim to learn a $d$-dimensional representation for every graph $\mathcal{G}^i$. We denote the number of nodes in $\mathcal{G}^i$ as $|\mathcal{G}^i|$ and the matrix of representations of all graphs as $X_{\mathbf{G}}\in \mathbf{R}^{n\times d}$.

\textbf{Semi-supervised Learning on Graphs.} In semi-supervised case, given a set of labeled graphs $\mathbf{G}^L=\{\mathcal{G}^1,\mathcal{G}^2,\dots,\mathcal{G}^{|\mathbf{G}^L|}\}$ with corresponing label $\{y^1,y^2,\dots,y^{|\mathbf{G}^L|}\}$ and a set of unlabeled graphs $\mathbf{G}^U=\{\mathcal{G}^1,\mathcal{G}^2,\dots\mathcal{G}^{|\mathbf{G}^U|}\}$, we aim to train a model to predict the labels of unseen graphs. In most cases $|\mathbf{G}^L|\ll|\mathbf{G}^U|$.

\textbf{Notations on Graphs.} We denote a graph $\mathcal{G}=(V,E,A)$, where $V=\{v_1,v_2,...,v_{\mid V\mid}\}$ is the set of nodes, $E=\{e_{ij}=(v_i,v_j)|v_i,v_j\in V\}$ is the set of edges, and $A\in\mathbf{R}^{|V|\times|V|}$ is the adjacency matrix with $A_{ij}>0$ if $e_{ij}\in E$ and $A_{ij}=0$ if $e_{ij}\notin E$. We also have node attributes $X_V\in\mathbf{R}^{|V|\times D_{V}}$ and edge attributes $X_E\in\mathbf{R}^{|E|\times D_{E}}$. 

In GNN, the general update formula of the $k$ th layer can be described as:
\begin{equation}\label{GNN agg}
X^{(k)}=h_{\Theta}^{(k)}\left(X^{(k-1)}, f_{\Phi}^{(k)}\left(A\cdot X^{(k-1)}\right)\right)
\end{equation}
where $h_{\Theta}^{(k)}$, $f_{\Phi}^{(k)}$ denote a neural network such as multilayer perceptron(MLP) and aggregation function for neighbors of nodes within the graph at layer $k$. Assume we pass the initial representations of nodes $X^{(0)}$ through $L$ layers of GNN, we can get a list of node representations: $\left\{X^{(1)}, X^{(2)}, ..., X^{(L)}\right\}$.

When we get the new reconstructed graph by 
subgraphs generation, which will be described in Section~\ref{SAS}, we denote subgraphs sampled from $\mathcal{G}$ as $\mathcal{G}_1, \mathcal{G}_1, ..., \mathcal{G}_{S}$, where $S$ is the number of subgraphs.

\subsection{Node-Agg Stage}\label{NAS}
In this stage, we get an enhanced node representation by the proposed \textbf{attribute-conv}, which can aggregate different perspectives of information into corresponding nodes. As is well known, GNN aggregates information from nodes by the connection relationship from the adjacency matrix. As a consequence, the quality of node representations directly affects the quality of the final graph representations obtained from GNN. Collected graph data usually have abundant information such as in quantum chemistry~\cite{gilmer2017neural}, node and edge attributes are both available for our study so we can use a lightweight convolution kernel to fuse node and edge attributes for downstream feedforward process. What's more, not limited to nodes and edges, anything useful information for our analysis can be aggregated into nodes such as specific local structure information.

Formally, assume we have $N$ kinds of node attributes:
$
X_1, X_2,...,X_N
$
Formally, assume we have $N$ kinds of node attributes:
where $X_i\in\mathbf{R}^{|V|\times N_i}$ where $N_i$ is the dimension of the $i\text{-}th$ kind of node attributes. For simplicity, we consider the case where there is the only node attributes $X_V\in\mathbf{R}^{|V|\times D_V}$ and edge attributes $X_E\in\mathbf{R}^{|E|\times D_E}$. The General case can be extended according to the specific setting. We first transform them into the same dimension embeddings by two MLPs:
\begin{align}\label{2}
X_{V}^{(0)}&=\textbf{MLP}_{\textbf{V}}(X_V)\in\mathbf{R}^{|V|\times d},\\
X_{E}^{(0)}&=\textbf{AGG}\left(\textbf{MLP}_{\textbf{E}}(X_E)\right)\in\mathbf{R}^{|V|\times d}.
\end{align}
\textbf{AGG} means aggregating edge embeddings into corresponding nodes. After getting the initial embeddings of nodes and edges: $X_{V}^{(0)}, X_{E}^{(0)}\in\mathbf{R}^{|V|\times d}$ , we use a $(2,1)$ size convolution kernel which we call \textbf{attribute-conv}, to squeeze each perspective of embeddings to one channel:
\begin{equation}\label{4}
X^{(0)} = \textbf{attribute-conv}([X_{V}^{(0)};X_E^{(0)}])\in\mathbf{R}^{|V|\times d},
\end{equation}
where $[\cdot\ ;...\ ;\ \cdot]$ means the operation of concatenating vectors. After all, we get the node embeddings used for the initial input of GNNs $X^{(0)}$. 
\subsection{Layer-Agg Stage}\label{LAS}
GNN aggregates the information of multi-hop neighbors successively as the number of GNN layers increases and the information contained in the representations of different hops will gradually change from locally to globally. 
That is, after getting the initial node embeddings $X^{(0)}$ and the feedforward of $L$ GNNs layers by Eq.~\ref{GNN agg}, we can get a list of node representations: $(X^{(1)}, X^{(2)}, ..., X^{(L)})$. $X^{(L)}$ is generally used but it will inevitably lose some distinguishable node information. As a consequence, we use a $(L,1)$ convolution kernel which we call \textbf{layer-conv} to aggregate the node representations of different scales so that local and global information can be combined organically:
\begin{equation}\label{5}
X_{\mathcal{G}} = \textbf{layer-conv}([X^{(1)}; X^{(2)}; ...; X^{(L)}])\in\mathbf{R}^{|V|\times d},
\end{equation}
After all, we get the final node embeddings which contain different levels information. Then we can get the whole graph represention by a readout function:
\begin{equation}\label{ori}
h(\mathcal{G}) = \textbf{READOUT}(X_{\mathcal{G}})\in\mathbf{R}^{d}.
\end{equation}

\subsection{Subgraph-Agg Stage}\label{SAS}

Since the nodes and graph representations do not express the same level of information, we find that maximizing the mutual information between nodes and graph representation is not good enough to achieve the purpose of graph representation to express more information. Therefore, we propose an ensemble learning-like subgraph method. First, we build an autoregressive model to generate several subgraphs from the original graph and then inspired by ensemble learning, we assemble these subgraphs into a reconstructed graph $\mathbf{G}^{\text{rec}}$. In this way, we can learn the graph representation, $h(\mathcal{G})$ from Eq.~\ref{5}, by maximizing the mutual information between two graphs representations in the same level, which can be described as:
\begin{align}\label{mi obj}
 \max \mathcal{I}(h(\mathcal{G});\tilde{h}(\mathcal{G})).
\end{align} 
where $\tilde{h}(\mathcal{G})$ denotes the representation of $\mathbf{G}^{\text{rec}}$.

%
%

In the following, we introduce the subgraph generation and graph reconstruction method respectively.

\textbf{Auto-Regressive Subgraph Generation.} We denote the original graph and the reconstructed graph as $\mathbf{G}\ \text{and}\  \mathbf{G}^{\text{rec}}=\{\mathcal{G}_{i},i=1,2,...,S\}$ respectively, where $\mathcal{G}_{i}$ is the $i\text{-}th$ subgraph and $S$ is the number of subgraphs. 
Equivalent to Eq.~\ref{mi obj}, we have:
\begin{align}
\max \mathcal{I}(\mathbf{G};\mathbf{G}^{\text{rec}})&=\mathbb{E}_{\mathbf{G}}\big[\mathbf{KL}(p(\mathbf{G}^{\text{rec}}|\mathbf{G})||p(\mathbf{G}^{\text{rec}}))\big],\nonumber
\end{align}
where $\mathbf{KL}(\ \cdot\ ||\ \cdot\ )$ means Kullback-Leibler Divergence. For the generation of subgraphs, we choose an auto-regressive model to achieve it, which can be written as:
\begin{align}
p(\mathbf{G}^{\text{rec}}|\mathbf{G})&=p(\{\mathcal{G}_{i},i=1,2,...,S\}|\mathbf{G})\nonumber\\
&=\prod_{i=2}^{S}p(\mathcal{G}_{i}|\mathbf{G},\mathcal{G}_{1},...,\mathcal{G}_{i-1}).
\end{align}


From this perspective of view, the generated subgraphs $\{\mathcal{G}_{i},i=1,2,...,S\}$ can obtain at least all the nodes and structural information of the original graph. Then with the guidance of our objective function Eq~\ref{mi obj}, the learned graph representation $h(\mathcal{G})$ can express the graph well enough.
Moreover, if the number of subgraphs $S$ is determined, all the above processes can be trained in an end-to-end way. In the next part, we will describe how to represent the original graph $\mathbf{G}$ and the reconstructed graph $\mathbf{G}^{\text{rec}}$.

\textbf{Graph Reconstruction.} 
We generate a group of subgraphs $\{\mathcal{G}_{i},i=1,2,...,S\}$ by an aforementioned auto-regressive way. We denote the node representations of each subgraph $\mathcal{G}_i$ as $X_{\mathcal{G}_i}$, $ i=1,2,...,S$,
then after a readout function, we can get the graph level representation of $\mathcal{G}_i$:
\begin{equation}\label{7}
h(\mathcal{G}_i) = \textbf{READOUT}(X_{\mathcal{G}_i})\in\mathbf{R}^{d},\ \ i=1,2,...,S.
\end{equation}

Like ensemble learning, we aggregate these $S$ 'weak' subgraphs in a 'strong' reconstructed graph, through a function called \textbf{subgraph-conv}. Specifically, we use a $(S,1)$ size convolution kernel to achieve the aggregation and we treat the resulting representation as the representation of the reconstructed graph $\mathbf{G}^{\text{rec}}$:
\begin{equation}\label{pos}
\tilde{h}(\mathcal{G}) = \textbf{subgraph-conv}([h(\mathcal{G}_1);h(\mathcal{G}_2);...;h(\mathcal{G}_S)])\in\mathbf{R}^{d},
\end{equation}

Overall, we generate several subgraphs to express the original graph more properly, and the proposed aggregation way ensures the information in $\mathbf{G}^{\text{rec}}$ is meaningful to guide the graph representation $h(\mathcal{G})$. And the graph-graph objective Eq.~\ref{mi obj} empirically leads to lower variance than node-graph constrain.


  
We also find that some subgraph generation methods in the latest researches, such as~\cite{li2020graph} and~\cite{hassani2020contrastive}. The start point of the sampling subgraph in the first work is multiscale graph representations~\cite{fu2019learning, gao2019graph, lee2019self,liao2019lanczosnet}. They get a subgraph by maximizing the mutual information between the whole graph and subgraph. The second work's idea is based on multi-view representation learning~\cite{tian2019contrastive,bachman2019learning} and they get subgraphs by Graph Diffusion Networks~\cite{klicpera2019diffusion}. However in our work, we get several subgraphs in an auto-regressive way, and the previous work can be seen as a special case of our universial framework.
 

  
\section{Implementation details}
 
\subsection{Subgraph Generation}\label{SG}
For the auto-regressive paradigm of subgraph generation described in Section~\ref{SAS}, we have simplified this process for the convenience of practical realization. We propose two approaches in our implementation: Tree-split generation and Multi-head generation. Both approaches are based on a basic operator, we first describe the basic operator and then introducing two specific implementations.

\textbf{Basic Operator.} After getting the node representations $X_{\mathcal{G}}\in\mathbf{R}^{|V|\times d}$ in Eq.~\ref{5}, we make a linear transformation on it by a learnable matrix $W\in\mathbf{R}^{d\times 2}$, and then we obtain a probability matrix $P=(P_1,...,P_{|V|})^{\mathbf{T}}\in\mathbf{R}^{|V|\times 2}$ by a row-wise softmax function. Formally, we get $P$ by:
\begin{equation}\label{P}
P = \textbf{Softmax}(X_{\mathcal{G}}\cdot W)\in\mathbf{R}^{|V|\times 2},
\end{equation}
and we describe the process of obtaining the probability matrix $P$ as a basic operator. And $p_{ij}$ denotes the probability that the $i$-th node is in the $j$-th subgraph, thus matrix $P$ divides the original graph into two subgraphs.

\begin{figure}[ht]
	\vskip 0.2in
	\begin{center}
		\centerline{\includegraphics[width=0.85\columnwidth]{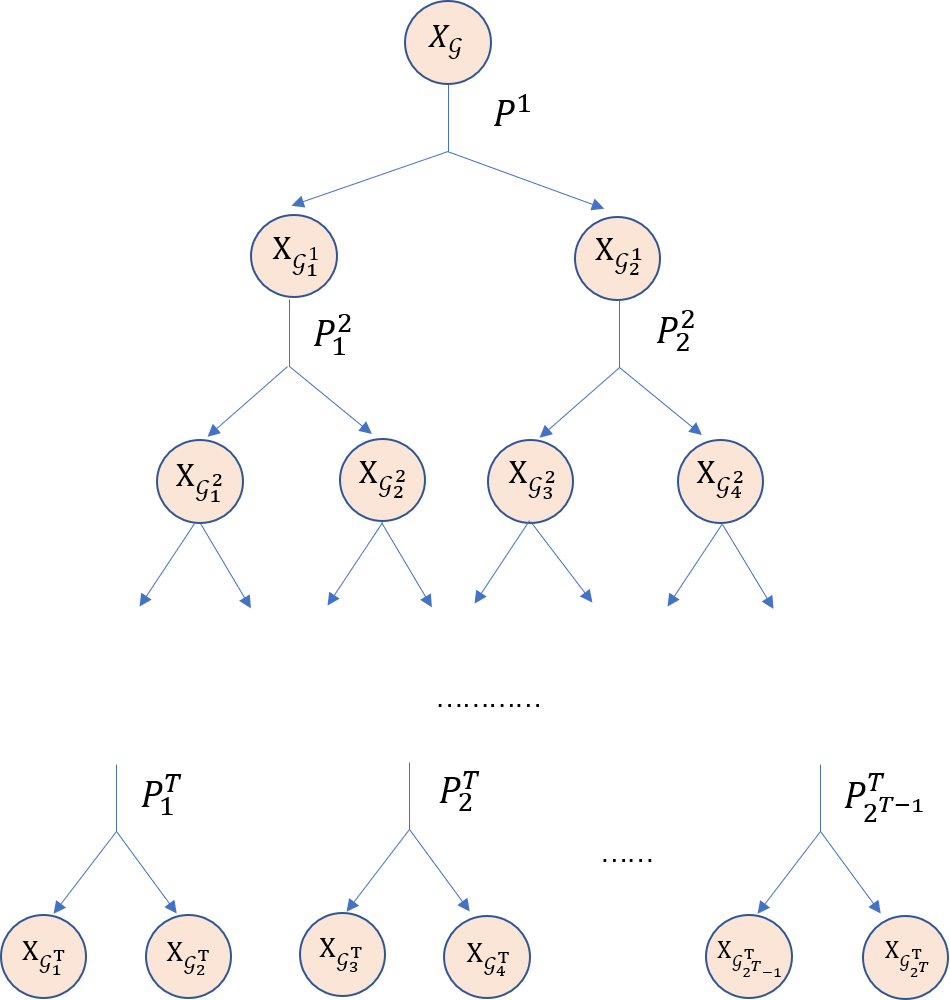}}
		\caption{Illustration of the Tree-split generation method.}
		\label{TS}
	\end{center}
	\vskip -0.2in
\end{figure}

\textbf{Tree-split Generator.}
In the Tree-split generator method, we recursively utilize the basic operator to the newly generated subgraph. We can see this kind of generation as the process of splitting a binary tree. At each non-leaf node, we execute the basic operator to get a probability matrix $P$ which represents the partition on the subgraph. And after $T$ rounds splits, we get $S = 2^T$ subgraphs $\{X_{\mathcal{G}^T_1},...,X_{\mathcal{G}^T_{2^T}}\}$, just as shown in Fig.~\ref{TS}. This method fits with the auto-regressive approach partially.

\textbf{Multi-head Generator.} Similar to the multi-head attention mechanism, we can introduce $S$ learnable matrices $\{W^1,...,W^S\}$ to execute the basic operator and get $S$ probability matrices $\{P^1,...,P^S\}$. We select the $i\text{-}th$ subgraph by the first column of $P^i$: if $P^i_{j,1}\ge \frac{1}{2}$, then the $j\text{-}th$ node is in the subgraph $i$, else not. In this way, we can select $S$ subgraphs in parallel but break the rule of auto-regressive generation. But generally, we can also see this kind of generation method in an auto-regressive way:
\begin{align}
p(\mathbf{G}^{\text{rec}}|\mathbf{G})&=p(\{\mathcal{G}_{i},i=1,2,...,S\}|\mathbf{G})\nonumber\\
&\approx\prod_{i=2}^{S}p(\mathcal{G}_{i}|\mathbf{G})
\end{align}
by assuming that all subgraphs are conditionally independent concerning the original graph.

\subsection{Loss Function}\label{LF}
We consider our model in both unsupervised and semi-supervised cases. Let $\phi$ denote the set of parameters of $L$-layers graph neural network and subgraph generation network.

In the unsupervised case, our model networks are parameterized by $\phi$ together and we seek to obtain a graph representation by maximizing the mutual information between the global graph representation $h_\phi(\mathcal{G})$ in Eq.~\ref{ori} and the reconstructed graph representation $\tilde{h}_{\phi}(\mathcal{G})$ in Eq.~\ref{pos}:
\begin{align}
\max \mathcal{L}_{\phi,\omega}^{uns}\triangleq&\sum_{\mathcal{G}\in\mathbf{G}}\frac{1}{|\mathbf{G}|}\mathcal{L}_{\phi,\omega}^{uns}(\mathcal{G})\nonumber\\
=&\sum_{\mathcal{G}\in\mathbf{G}}\frac{1}{|\mathbf{G}|}I_{\phi,\omega}(h_\phi(\mathcal{G});\tilde{h}_\phi(\mathcal{G}))
\label{uns_ori}
\end{align}

$I_{\phi,\omega}(h_\phi(\mathcal{G});\tilde{h}_\phi(\mathcal{G}))$ is the mutual information estimator modeled by dicriminator $T_{\omega}: h_{\phi}(\mathcal{G})\times \tilde{h}_{\phi}(\mathcal{G})\rightarrow \mathbf{R}$ that scores the representation pairs (positive pairs have higher scores). Here we can use Jensen-Shannon Divergence (JSD)~\cite{nowozin2016f} estimator: 
\begin{align}
I_{\phi,\omega}^{JSD}(h_{\phi}&(\mathcal{G});\tilde{h}_{\phi}(\mathcal{G}))\nonumber\\
=&\mathbf{E}_{P}(-sp(-T_{\omega}(h_{\phi}(\mathcal{G}),\tilde{h}_{\phi}(\mathcal{G})))\nonumber \\
&-\mathbf{E}_{Q}(sp(T_{\omega}(h_{\phi}(\mathcal{G}),\tilde{h}_{\phi}(\mathcal{G}))))
\label{mi}
\end{align}

$P=p(h_{\phi}(\mathcal{G}),\tilde{h}_{\phi}(\mathcal{G}))$ is the joint distribution of global graph representaion and reconstructed graph representation, while
$Q=p(h_{\phi}(\mathcal{G}))p(\tilde{h}_{\phi}(\mathcal{G}))$ denotes the multiplication of marginal distributions of two representations. In constrasive learning, $Q$ indicates the distribution of positive pairs and $P$ indicates the distribution of negative pairs. And $sp(x)=\log(1+e^x)$ is the softplus function.

Besides the JSD estimator, we can also achieve it by using the estimator based on Donsker-Varadhan representation (DV)~\cite{donsker1983asymptotic},
\setlength\arraycolsep{0pt}
\begin{align}
I_{\phi,\omega}^{DV}(h_{\phi}&(\mathcal{G});\tilde{h}_{\phi}(\mathcal{G}))\nonumber\\
=&\mathbf{E}_{P}(T_{\omega}(h_{\phi}(\mathcal{G}),\tilde{h}_{\phi}(\mathcal{G})))\nonumber\\
&-\log\mathbf{E}_{Q}(e^{T_{\omega}(h_{\phi}(\mathcal{G}),\tilde{h}_{\phi}(\mathcal{G}))})
\end{align}

In general, there are two ways to obtain negative samples, the first using different graphs in the batch, and the other using the corruption function to get negative samples from the original graph.

Here, we use a Head-Tail negative pair samples used for contrastive method. For graph $\mathcal{G}^i$, we use the global representation of different graphs, $h_\phi(\mathcal{G}^j),j\ne i$ ,in the dataset to generate Tail negative pair samples ($h_{\phi}(\mathcal{G}^j),\tilde{h}_{\phi}(\mathcal{G}^i)$). Besides, we can get a negative graph $\mathcal{\hat{G}}^i$ specifically for $\mathcal{G}^i$, by shuffling the node embeddings as:
\begin{equation}\label{9}
\hat{X}^{(0)}=\textbf{Permute}(X^{(0)})\in\mathbf{R}^{|V|\times d},
\end{equation}
and then get the global graph representation $h_{\phi}(\mathcal{\hat{G}}^i)$. Such that we can obtain the Head negative pair samples ($h_{\phi}(\mathcal{\hat{G}}^i),\tilde{h}_{\phi}(\mathcal{G}^i)$), uniquely for graph $\mathcal{G}^i$.

In this way, our model yields a two-part negative sample term, therefore, the distribution $Q$ in Eq\ref{mi} can be separated into two parts:
\begin{align}
\mathcal{L}^{uns}_{\phi,\omega}(\mathcal{G})=&\mathbf{E}_{p(h_{\phi}(\mathcal{G}),\tilde{h}_{\phi}(\mathcal{G}))}(-sp(-T_{\omega}(h_{\phi}(\mathcal{G}),\tilde{h}_{\phi}(\mathcal{G}))) \nonumber\\
&-\mathbf{E}_{p(h_{\phi}(\mathcal{G}))p(\tilde{h}_{\phi}(\mathcal{G}))}(sp(T_{\omega}(h_{\phi}(\mathcal{G}),\tilde{h}_{\phi}(\mathcal{G})))\nonumber\\
&-\mathbf{E}_{p(h_{\phi}(\mathcal{\hat{G}}),\tilde{h}_{\phi}(\mathcal{G}))}(sp(T_{\omega}(h_{\phi}(\mathcal{\hat{G}}),\tilde{h}_{\phi}(\mathcal{G})))
\label{uns}
\end{align}

In the semi-supervised case, we also use the labeled data to make predictions and get the cross-entropy loss $\mathcal{L}^{sup}_{\phi,\theta}(y_\phi(\mathcal{G}^i);y^i)=\mathbb{E}_{p_\phi(\mathcal{G}^i;h(\mathcal{G}^i))}\log p_\theta(y|h(\mathcal{G}^i))$, where $\theta$ denotes the parameters of classfier network. And for unlabeled data, we make the same constrain as unsupervised case. Thus we can describe the semi-supervised model as:
\begin{small}
	\begin{align}
	\max \mathcal{L}^{semi}_{\phi,\omega,\theta}=\sum_{i=1}^{|\mathbf{G}^L|}\mathcal{L}^{sup}_{\phi,\theta}(y_\phi(\mathcal{G}^i);y^i)+\lambda\sum_{j=1}^{|\mathbf{G}^L|+|\mathbf{G}^U|}\mathcal{L}^{uns}_{\phi,\omega}(\mathcal{G}^j)
	\label{semi}
	\end{align}
\end{small}
the hyper-parameter $\lambda$ balances the supervised and unsupervised loss.

\begin{figure*}[ht]
	\vskip 0.2in
	\begin{center}
		\centerline{\includegraphics[width=1.8\columnwidth]{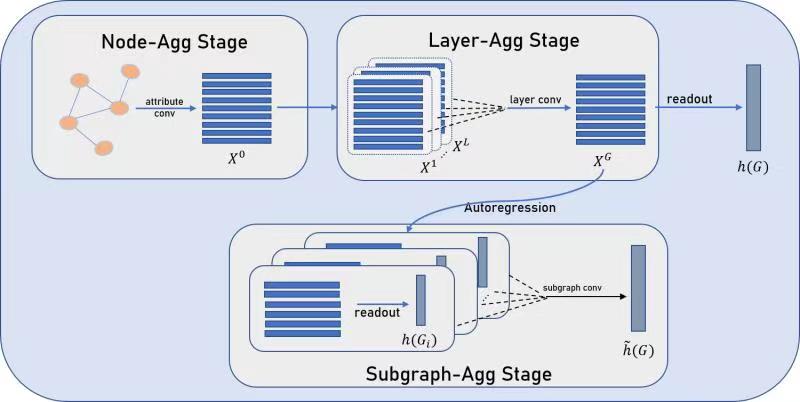}}
		\caption{Illustration of our proposed method. There are three stages in the feedforward process: Node-Agg stage, Layer-Agg stage, and Subgraph-Agg stage. In Node-Agg stage, we aggregate original attributes by \textbf{attribute-conv} to get the initial input $X^0$ for GNNs. In Layer-Agg stage, we get the global node embedding $X^G$ by using \textbf{layer-conv} to aggregate different scales node embeddings. On one hand, we can obtain the global graph embedding $h(G)$ by a readout function. On the other hand, we can generate several subgraphs in an auto-regressive way and then get a reconstructed graph embedding $\tilde{h}(G)$ by \textbf{subgraph-conv}. After all, we get two graph embeddings $h(G)$ and $\tilde{h}(G)$ which are at the same level. During training period, we can guide $h(G)$ by $\tilde{h}(G)$ by contrastive learning. And during the testing period, we use the trained model to get $h(G)$ for downstream tasks.}
		\label{architecture}
	\end{center}
	\vskip -0.2in
\end{figure*}

Fig~\ref{architecture} shows the model architecture and the process of parameter updating for one batch of data are summarized in Algorithm~\ref{alg:1}.
\begin{algorithm}[h]
	\caption{}
	\label{alg:1}
	\begin{algorithmic}[1]
		\STATE \textbf{Input:} A batch of graph data $\mathbf{G}=\{\mathcal{G}^1,\mathcal{G}^2,\dots\}$ with corresponding  node attributes $\{X_{V^1},X_{V^2},...\}$, edge attributes $\{X_{E^1},X_{E^2},...\}$, hyperparameter $\lambda$ for semi-supervised case
		\STATE \textbf{Networks Initialization}:\\
		\ \ \textbf{GNNs}: $\phi=\{h_{\Theta}^{(k)}, f_{\Phi}^{(k)},k=1,...,L\}$;\\
		\ \ \textbf{attribute-conv}: $(2,1)$ size CNN;\\
		\ \ \textbf{layer-conv}: $(L,1)$ size CNN;\\
		\ \ \textbf{subgraph-conv}: $(S,1)$ size CNN;\\
		\ \ \textbf{Discriminator}: $T_{\omega}: h_{\phi}(\mathcal{G})\times \tilde{h}_{\phi}(\mathcal{G})\rightarrow \mathbf{R}$\\
		\ \ $\textbf{MLP}_{\textbf{V}}$, $\textbf{MLP}_{\textbf{E}}$
		\STATE Get initial inputs of GNNs $\{X^{(0)}_1,X^{(0)}_2,...\}$ according Eq.~\ref{2}$-$\ref{4}
		\STATE Get graph representations $\{h(\mathcal{G}^1),h(\mathcal{G}^2),...\}$ according to Eq.~\ref{5}$-$\ref{ori}; 
		\STATE Get head negative samples $\{h(\hat{\mathcal{G}^1}),h(\hat{\mathcal{G}^2}),...\}$ by Permute function 
		Eq.~\ref{9};
		\FOR{each $\mathcal{G}^i\in \mathbf{G}$}
		\STATE Generating $S$ subgraphs $\{\mathcal{G}_1^i, ..., \mathcal{G}_{S}^i\}$ with corresponding global node representations $\{X_{\mathcal{G}_1^i},X_{\mathcal{G}_2^i},...X_{\mathcal{G}_{S}^i}\}$ from $X_{\mathcal{G}^i}$ by Tree-split generation or Multi-head generation, and getting positive samples $\tilde{h}(\mathcal{G}^i)$ according to Eq.~\ref{7}$-$\ref{pos}
		\ENDFOR
		\IF{in unsupervised case}
		\STATE Optimize Eq.~\ref{uns} by collecting positive sample pairs $\{(h(\mathcal{G}^i),\tilde{h}(\mathcal{G}^i)),i=1,...,|\mathbf{G}|\}$, Head negative sample pairs $\{(h(\hat{\mathcal{G}^i}),\tilde{h}(\mathcal{G}^i)),i=1,...,|\mathbf{G}|\}$ and Tail negative sample pairs $\{(h(\mathcal{G}^i),h(\mathcal{G}^j),i\ne j\}$
		\ENDIF
		\IF{in semi-supervised case}
		\STATE Optimize Eq.~\ref{semi} where the construction of unsupervised term is the same as in an unsupervised case
		\ENDIF
	\end{algorithmic}
\end{algorithm}

\section{Experimental Results}
\subsection{Datasets}
\begin{table*}[t]
	\caption{Stastics of the datasets used in our experiments}
	\label{table:datasets}
	\setlength{\tabcolsep}{3.5pt}
		\vskip 0.05in
	\begin{center}
		\begin{small}
			\begin{sc}
				\begin{tabular}{c|c|c|c|c|c|c}
					\toprule
					\textbf{}                   & \textbf{MUTAG} & \textbf{PTC-MR} & \textbf{IMDB-BINARY} & \textbf{IMDB-MULTI} & \textbf{REDDIT-BINARY} & \textbf{REEDIT-M5K} \\ 
					\midrule
					\multicolumn{1}{l|}{\textbf{Graph Numbers}}    & 188            & 344              & 1000                 & 1500                & 2000                   & 4999                    \\
					
					\multicolumn{1}{l|}{\textbf{Class Numbers}}    & 2              & 2                & 2                    & 3                   & 2                      & 5                       \\
					\multicolumn{1}{l|}{\textbf{Avg. Edges}} & 19.79          & 14.69            &  193.06                &  65.93              & 497.75                & 508.52                  \\
					\multicolumn{1}{l|}{\textbf{Avg. Nodes}} & 17.93          & 14.29            & 19.77                & 13.00               & 429.63                 & 594.87                 
				\end{tabular}
			\end{sc}
		\end{small}
	\end{center}
	\vskip -0.2in
\end{table*}
\begin{table*}
	\caption{Unsupervised Results: The accuracy results of the kernel, supervised, and unsupervised models in downstream graph classification task.}
	\label{table:uns_results}
	\setlength{\tabcolsep}{3.5pt}
	\begin{center}
		\begin{small}
			\begin{sc}
				\scalebox{0.89}{
					\begin{tabular}{clccccccc}
						\toprule
						& \textbf{Methods}& \textbf{MUTAG}             &\textbf{ PTC-MR}           & \textbf{IMDB-B}       & \textbf{IMDB-M}        & \textbf{RDT-B}     & \textbf{RDT-M5K} \\ 
						\midrule
						\multirow{6}{*}{\begin{turn}{90}
								\textbf{Kernel}
						\end{turn}}&\textbf{SP}~\cite{borgwardt2005shortest}  & 85.2 $\pm$ 2.4 & 58.2 $\pm$ 2.4 & 55.6 $\pm$ 0.2 & 38.0 $\pm$ 0.3 & 64.1 $\pm$ 0.3 &39.6 $\pm$ 0.2                         \\
						&\textbf{GK}~\cite{shervashidze2009efficient}  & 81.7 $\pm$ 2.1 & 57.3 $\pm$ 1.4 & 65.9 $\pm$ 1.0 & 43.9 $\pm$ 0.4 & 77.3 $\pm$ 0.2 & 41.0 $\pm$ 0.2                     \\
						&\textbf{WL}~\cite{shervashidze2011weisfeiler}  & 80.7 $\pm$ 3.0 & 58.0 $\pm$ 0.5 & \textbf{72.3 $\pm$ 3.4} & \textbf{47.0 $\pm$ 0.5} & 68.8 $\pm$ 0.4 & \textbf{46.1 $\pm$ 0.2}                     \\
						&\textbf{RW}~\cite{gartner2003graph}  & 83.7 $\pm$ 1.5 & 57.9 $\pm$ 1.3 & 50.7 $\pm$ 0.3 & 34.7 $\pm$ 0.2 &             $-$                &             $-$             \\
						&\textbf{DGK}~\cite{yanardag2015deep} & 87.4 $\pm$ 2.7 & 60.1 $\pm$ 2.6 & 67.0 $\pm$ 0.6 & 44.6 $\pm$ 0.5 & \textbf{78.0 $\pm$ 0.4} &41.3 $\pm$ 0.2 \\
						&\textbf{MLG}~\cite{kondor2016multiscale} & \textbf{87.9 $\pm$ 1.6} & \textbf{63.3 $\pm$ 1.5} & 66.6 $\pm$ 0.3 & 41.2 $\pm$ 0.0   &$-$&$-$                     \\ 
						\midrule
						\multirow{5}{*}{\begin{turn}{90}
								\textbf{Supervised}
						\end{turn}}
						&\textbf{GraphSage}~\cite{hamilton2017inductive}  & 81.1 $\pm$ 7.6 & 63.9 $\pm$ 7.7 & 72.3 $\pm$ 5.3 & 50.9 $\pm$ 2.2 &  &                   \\
						&\textbf{GCN}~\cite{kipf2016semi}  & 85.6 $\pm$ 5.8 & 64.2 $\pm$ 4.3 & 74.0 $\pm$ 3.4 & 51.9 $\pm$ 3.8 & 50.0 $\pm$ 0.0 &                     \\
						&\textbf{GIN-0}~\cite{xu2018powerful}  & \textbf{89.4 $\pm$ 5.6} & 64.6 $\pm$ 7.0 &  {\textbf{75.1 $\pm$ 5.1}} &  {\textbf{52.3 $\pm$ 2.8}} &    {\textbf{92.5 $\pm$ 2.5}}    &    {\textbf{57.5 $\pm$ 1.5}}          \\
						&\textbf{GIN-$\epsilon$}~\cite{xu2018powerful} & 89.0 $\pm$ 6.0 & 63.7 $\pm$ 8.2 & 74.3 $\pm$ 5.1 & 52.1 $\pm$ 3.6 & 92.2 $\pm$ 2.3&     57.0 $\pm$ 1.7          \\
						&\textbf{GAT}~\cite{velivckovic2017graph} & \textbf{89.4 $\pm$ 6.1} & {\textbf{66.7 $\pm$ 5.1}} & 70.5 $\pm$ 2.3 & 47.8 $\pm$ 3.1  &85.2 $\pm$ 3.3 &               \\ 
						\midrule 
						\multirow{6}{*}{\begin{turn}{90}\textbf{unsupervised}\end{turn}}&\textbf{Node2Vec}~\cite{grover2016node2vec}     & 72.6 $\pm$ 10.2  & 58.6 $\pm$ 8.0     & $-$             & $-$                    & $-$                       &     $-$                     \\
						&\textbf{Sub2Vec}~\cite{adhikari2018sub2vec}       & 61.6 $\pm$ 15.8 & 60.0 $\pm$ 6.4   & 55.3 $\pm$ 1.5 & 36.7 $\pm$ 0.8        & 71.5 $\pm$ 0.4           &  36.7 $\pm$ 0.4                      \\
						&\textbf{Graph2Vec}~\cite{narayanan2017graph2vec}     & 83.2 $\pm$ 9.6   & 60.2 $\pm$ 6.9     & 71.1 $\pm$ 0.5 & 50.4 $\pm$ 0.9        & 75.8 $\pm$ 1.0           & 47.9 $\pm$ 0.3                        \\
						&\textbf{InfoGraph}~\cite{sun2019infograph}     & 89.0 $\pm$ 1.1   & 61.7 $\pm$ 1.4     & 73.0 $\pm$ 0.9 & 49.7 $\pm$ 0.5        & 82.5 $\pm$ 1.4           & 53.5 $\pm$ 1.1                       \\
						&\cite{hassani2020contrastive}              & 89.2 $\pm$ 1.1  & 62.5 $\pm$ 1.7   & \textbf{74.2 $\pm$ 0.7} & \textbf{51.2 $\pm$ 0.5}        & 84.5 $\pm$ 0.6          &      $-$                   \\          
						&\textbf{OURS(MH)}  &  91.7 $\pm$ 1.1          & 64.6 $\pm$ 0.6              &     73.2 $\pm$ 0.7               &          50.6 $\pm$ 0.4         &   \textbf{91.3 $\pm$ 0.6}        &  55.0 $\pm$ 0.5 \\
						&\textbf{OURS(TS)}  &   {\textbf{91.8 $\pm$0.5}}              &  \textbf{65.8 $\pm$ 1.3}              &     73.3 $\pm$ 0.5               &          50.5 $\pm$ 0.3         &  90.5 $\pm$ 0.2       &  \textbf{55.2 $\pm$ 0.3} \\
						\bottomrule
						
				\end{tabular}}
			\end{sc}
		\end{small}
	\end{center}
	\vskip -0.2in
\end{table*}

\begin{table*}[htp!]
	\caption{Semi-Supervised Results: The top table is the results of our model, and the bottom is the results of InfoGraph. The first rows in the top and bottom tables show the mean absolute error (MAE) of the supervised model, and the rest rows show the error ratio (concerning supervised result) of semi-supervised and Mean-Teacher model (lower scores indicate better performance than supervised model).}
	\setlength{\tabcolsep}{3.5pt}
	\vskip -0.1in
	\begin{center}
		\begin{small}
			\begin{sc}
				\scalebox{0.87}{
				\begin{tabular}{cl|cccccccccccc}
					\toprule
				
					& ~ & Mu(0) & Alpha(1) & HOMO(2) & LUMO(3) & Gap(4) & R2(5) & ZPVE(6) & U0(7) & U(8) & H(9) & G(10) & Cv(11)\\
					\midrule
					\multirow{1}{*}
					&MAE (Ours)  & 0.2581      & 0.5358      & 0.1589      & 0.1572      & 0.2312      & 4.0730       & 0.0099   & 6.9030       & 7.2901   & 5.4350      & 6.6210       & 0.1988      \\
					\midrule
					\multirow{3}{*}
					&Mean-Teachers& 0.98   & \textbf{0.93}   & 1.03   &0.97   & \textbf{0.97}   & \textbf{0.47}  & \textbf{0.80}      & \textbf{0.61}  & 0.82      & 1.09  & 0.76  & \textbf{0.85}   \\
					&Ours (TS) & 0.89   & 0.99
				   & \textbf{0.96}   & \textbf{0.93} & 0.99 & 0.79 & 0.91  & 0.52 & \textbf{0.54} &\textbf{0.90}  & 0.75& 0.90 \\
					&Ours (MH) & \textbf{0.82}   & 0.96 & 0.97   & 0.94  & 0.99   & 0.74  & 0.93  & 0.80  &0.87     &0.91  & \textbf{0.64} & 0.89
	                         \\
					\midrule
					\\
					\midrule
					\multirow{1}{*}
					&MAE (InfoGraph)     & 0.2216      & 0.5175      & 0.1577      & 0.1500        & 0.2317      & 3.3290       & 0.0104     & 6.0158      & 5.1360       & 5.2770       & 5.4383      & 0.2026      \\
					\midrule
					&Mean-Teacher     & 1.06        & \textbf{0.98} & \textbf{0.94} & 0.97          & 0.98          & \textbf{0.57} & \textbf{0.67} & \textbf{0.68} & \textbf{0.75} & \textbf{0.85} & \textbf{0.77} & \textbf{0.83} \\
					&InfoGraph  & 1.05          & 1.02           & 0.98          & \textbf{0.96} & \textbf{0.97} & 0.88          & 0.84          & 1.03         & 1.63          & 1.67          & 1.23           & 0.84          \\
					&InfoGraph* & \textbf{0.98} & 1.12          & 1.01       & 0.99          & 0.98           & 1.07          & 0.91          & 1.07          & 1.05          & 1.11          & 1.18         & 0.99  \\
					
					\bottomrule
				\end{tabular}}
			\end{sc}
		\end{small}
	\end{center}
	\vskip -0.2in
		\label{tab:semi_results}
\end{table*}

For unsupervised graph classification, we use the following datasets (statistics details shown in Table~\ref{table:datasets}) in TUDatasets~\cite{morris2020tudataset}: MUTAG~\cite{debnath1991structure,kriege2012subgraph} is a dataset of 188 mutagenic aromatic and heteroaromatic nitro compounds with 7 different discrete labels; PTC-MR~\cite{helma2001predictive,kriege2012subgraph} is a dataset of 344 chemical compounds that reports the carcinogenicity for male and female rats and it has 19 discrete labels; IMDB-BINARY and IMDB-MULTI~\cite{yanardag2015deep} are movie collaboration datasets with 2 and 3 discrete labels respectively. Each graph corresponds to a network of relationships between actors, where nodes correspond to actors/actresses, and when two actors appear in the same movie, an edge is drawn between them; REDDIT-BINARY and REDDIT-MULTI-5K~\cite{yanardag2015deep} are balanced datasets, where each graph corresponds to an online discussion thread and the nodes correspond to users. An edge is drawn between two nodes if at least one of them replies to a comment of the other node. The task is to classify each graph to the community it belongs to. 

For semi-supervised graph classification, we use QM9 dataset~\cite{wu2018moleculenet} which consists of about 130,000 molecules with 19 regression targets.
\subsection{Baselines}
In unsupervised case, we compare our method with 6 state of the art graph kernels: Random Walk (RW)~\cite{gartner2003graph}, Shortest Path Kernel (SP)~\cite{borgwardt2005shortest}, Graphlet Kernel (GK)~\cite{shervashidze2009efficient}, Weisfelier-Lehman Sub-tree Kernel (WL)~\cite{shervashidze2011weisfeiler}, Deep Graph Kernels (DGK)~\cite{yanardag2015deep}, and Multi-Scale Laplacian Kernel (MLG)~\cite{kondor2016multiscale} from InfoGraph. Besides comparing with graph kernel methods, we also compare our results with graph-level representation learning methods: Node2vec~\cite{grover2016node2vec}, Sub2vec~\cite{adhikari2018sub2vec}, Graph2vec~\cite{narayanan2017graph2vec}, InfoGraph~\cite{sun2019infograph} and~\cite{hassani2020contrastive}. We also present some results of supervised models like GraphSage~\cite{hamilton2017inductive}, Graph Convolution Network (GCN)~\cite{kipf2016semi}, GIN~\cite{xu2018powerful} and Graph Attention Network  (GAT)~\cite{velivckovic2017graph} from~\cite{hassani2020contrastive}.

\subsection{Experiment Configuration}\label{exp config}
 For unsupervised learning experiments, we evaluate our method for graph classification tasks and adopt the same procedure as InfoGraph~\cite{sun2019infograph}, use 10-fold cross-validation accuracy to report the classification performance. Experiments are repeated 7 times, the maximum and minimum values were removed, and then the average was taken. The classification accuracies are computed using LIBSVM~\cite{chang2011libsvm}, and the parameter $C$ was selected from $\{10^{-3},10^{-2},...,10^2,10^3\}$.

For semi-supervised learning experiments, we use the QM9 dataset and separate the data of each target in the dataset in the following way: 5000 random chosen samples as labeled samples, another 10000 as validation samples, another 10000 as test samples, and the rest as unlabeled training samples. We use the same way of dataset split when running the supervised model and the semi-supervised model. The validation set is used to do model selection and we report results on the test set. All targets were normalized to have mean 0 and variance 1. We utilize the mean square loss function to optimize our model and evaluate the mean absolute error.

\subsection{Model Configuration}\label{model config}
For the unsupervised experiments, we use GIN~\cite{xu2018powerful} as our base model which is on the same starting line with InfoGraph~\cite{sun2019infograph}. We use node degree as initial node attribute when datasets don't carry initial node features and there is no \textbf{attribute-conv} when lacking edge attributes. All hidden dimensions are set to 128, the batch size is 128, and the number of GNN layers is 4 so the \textbf{layer-conv} kernel size is $(4,1)$.  We empirically choose the number of subgraphs $S\in\{2, 4, 8\}$ in both subgraph generation method, so the \textbf{subgraph-conv} kernel size is $(S,1)$ corresponding to different settings. The initial learning rate is $10^{-3}$, the number of epochs is 100 and we report evaluation results by linear model per 5 epochs.

For semi-supervised experiments, the size of \textbf{attribute-conv} is $(2,1)$ since there exists node attributes and edge attributes, \textbf{layer-conv} size is $(3,1)$, \textbf{subgraph-conv} size is $(2,1)$ or $(4,1)$ in both subgraph generation methods and the number of set2set computations is set to 3. The classifier is a two-layers full-connection structure with ReLU activation function after the first linear layer. The initial learning rate is $10^{-3}$, training epoch number is 500 and the weight decay is 0. The hyper-parameter $\lambda$ in semi-supervised loss is $10^{-3}$. 

Both in the unsupervised and semi-supervised model, we directly compute the dot product of the graph representations and the reconstructed graph representations obtained by the model to calculate the discriminator scores of the sample pairs.
Models were trained using SGD with the Adam optimizer in both scenarios.
\begin{table*}[htp!]
	\caption{Effect of our proposed components with MH subgraph.}
	\label{table:ablation1}
	\setlength{\tabcolsep}{3.5pt}
	\begin{center}
		\begin{small}
			\begin{sc}
				\begin{tabular}{cl|ccccccc}
					\toprule
					& ~ &\textbf{mutag} & \textbf{ptc-mr} & \textbf{imdb-b} &\textbf{ imdb-m} & \textbf{reddit-b}\\
					\midrule
					\multirow{4}{*}
					&Infograph & 89.01$\pm$1.13   & 61.65$\pm$1.43     & 73.03$\pm$0.87 & 49.69$\pm$0.53        & 82.50$\pm$1.43 \\
					&Base(mh) & 90.26$\pm$0.42  & 63.16$\pm$1.25     & 73.03$\pm$0.54
					&   50.45$\pm$0.12   &89.35$\pm$0.63  &\\
					&Base+Neg(mh) & 91.11$\pm$1.29  &63.29$\pm$1.11     & 73.10$\pm$0.54
					&   50.49$\pm$0.30   & 89.41$\pm$0.32\\
					&Ours(mh) & 91.70$\pm$1.10          & 64.63$\pm$1.41              &     73.20$\pm$0.72               &          50.63$\pm$0.41         &   91.38$\pm$0.61     \\
				
					\bottomrule
				\end{tabular}
			\end{sc}
		\end{small}
	\end{center}
	\vskip -0.2in
\end{table*}
\begin{table*}[htp!]
	\caption{Effect of our proposed components with TS subgraph.}
	\label{table:ablation2}
	\setlength{\tabcolsep}{3.5pt}
	\begin{center}
		\begin{small}
			\begin{sc}
				\begin{tabular}{cl|ccccccc}
					\toprule
					& ~ &\textbf{mutag} & \textbf{ptc-mr} & \textbf{imdb-b} &\textbf{ imdb-m} & \textbf{reddit-b}\\
					\midrule
					\multirow{4}{*}
					&Infograph & 89.01$\pm$1.13   & 61.65$\pm$1.43     & 73.03$\pm$0.87 & 49.69$\pm$0.53        & 82.50$\pm$1.43 \\
					&Base(ts) & 90.31$\pm$0.92  & 62.94$\pm$0.76    & 73.04$\pm$0.48
					& 50.21$\pm$0.13     &88.50$\pm$0.85 \\
					&Base+Neg(ts) & 90.63$\pm$0.69  & 63.75$\pm$1.28     & 73.15$\pm$0.32
					&    50.43$\pm$0.43  & 88.80$\pm$1.10\\
						&Ours(ts) &  91.80$\pm$0.56      &  65.86$\pm$1.34              &     73.32$\pm$0.51               &          50.52$\pm$0.32         &  90.55$\pm$0.26  \\
					\bottomrule
				\end{tabular}
			\end{sc}
		\end{small}
	\end{center}
	\vskip -0.2in
\end{table*}
\subsection{Results}\label{results}
We evaluate the unsupervised model by downstream graph classification task, and the results are presented in Table~\ref{table:uns_results}. \textbf{OURS(TS)} means the subgraphs are generated by Tree-Split Generator~(TS) and \textbf{OURS(MH)} means the subgraphs are generated by Multi-Head Generator~(MH).
It is shown by the results that our method achieves state-of-the-art results in both unsupervised and kernel methods in MUTAG, PTC\_MR, REDDIT-BINARY, and REDDIT-MULTI-5K datasets, and even a competitive result with supervised models. In the IMDB-BINARY and IMDB-MULTI datasets, we are better than other methods except~\cite{hassani2020contrastive}.
Especially in REDDIT-BINARY datasets, we have $6.8\%$ accuracy ahead of the second place. We have different underlying graph neural network structures and our model gives more good explanations for the learned representation.

For semi-supervised tasks, we use the same settings and hyperparameters as InfoGraph~\cite{sun2019infograph}, and compare our method with the state-of-the-art method, Mean-Teacher~\cite{tarvainen2017mean} which is applicable for regression tasks. And we present results in Table~\ref{tab:semi_results}. We add the mutual information maximization objective to the purely supervised model and greatly improve the results. And we achieve better results in all 12 targets than the supervised model with both MH and TS generators. And compared with Mean-Teacher, our results are also competitive.

Moreover, we then reproduce the semi-supervised experiments with the code provided by InfoGraph for 5 times and take the best results, which are shown in the bottom part of Table~\ref{tab:semi_results}. It shows that InfoGraph does not perform well in our reproduce experiments. Furthermore, several results show that InfoGraph has negative effects in the semi-supervised case, however, all of our results have promotion compared to pure supervised case.

\subsection{Ablation Study}

In this section, we perform ablation experiments for our model to verify the validity of each component of our model, and we also make a comparison with InfoGraph~\cite{sun2019infograph}. Because several components are highly tangled, we set a based mode called \textbf{BASE} in Table~\ref{table:ablation1} and Tabel~\ref{table:ablation2}, with TS and MH subgraphs generator respectively,  which contains: \textbf{layer-conv} and generated subgraphs. In this case, we maximize the mutual information between the original graph and all subgraphs. \textbf{BASE+NEG} means adding head negative samples into \textbf{BASE}. Finally, \textbf{OURS}, which is our ultimate model, adds the \textbf{subgraph-conv} based on \textbf{BASE+NEG}. According to the recursive relationship, we can get a clear intuition on the effects of our main components.
Due to the computation and time cost, we only do ablation study on MUTAG, PTC-MR, REDDIT-BINARY, IMDB-BINARY, and IMDB-MULTI. Model configuration is same as in Section~\ref{exp config} and Section~\ref{model config}.

\textbf{Effects on Subgraphs and Layer-Conv.} BASE(MH) and BASE(TS) in the second rows of Table~\ref{table:ablation1} and~\ref{table:ablation2} add generated subgraphs and \textbf{layer-conv} based on InfoGraph. We can see that generated subgraphs and \textbf{layer-conv} can improve performance on most datasets compared with InfoGraph except for IMDB-BINARY. Especially in the REDDIT-BINARY dataset, these two components can increase by about $6\%$.

\textbf{Effects on Head-Tail Contrastive Sampling Method.} When we add head negative samples to \textbf{BASE}, performance on all datasets has improved in both generating methods.

\textbf{Effects on Aggregation Convolution Kernels.} The comparison of the third and last row demonstrates the addition of \textbf{subgraph-conv} significantly improves the results of our model, especially in MUTAG, PTC-MR, and REDDIT-BINARY datasets. And these above experiments also prove that our ultimate model is the best.

These ablation experiments demonstrate that each of our components can independently contribute to the model. Therefore, in future work, we can use these components to help any model to get further improvement.


\section{Conclusion and Future Work}
In this paper, we propose a self-supervised method to learn graph representations by maximizing the mutual information between the original graph and the reconstructed graph. 
To properly aggregate the information from the original graph, we utilize \textbf{attribute-conv} to aggregate the raw attributes, \textbf{layer-conv} to fuse different scales of information obtained from different layers of GNN and \textbf{subgraph-conv} to mix several generated subgraphs information. For the subgraph generation, we propose an auto-regressive method which can be seen as a universal framework to generate subgraphs in a learnable way. In particular, we introduce two specific subgraph generators: Tree-Split Generator and Multi-Head Generator. For making a better constraint on graph embeddings, we use a so-called \textbf{Head-Tail} contrastive sample construction to provide more negative samples which are beneficial for contrastive learning.
By all components above, we achieve state-of-the-art results compared to previous works by a big margin in several graph classification datasets.

For future work, we will dissect the influence of graph structure in the different datasets on our proposed components and improve the performance on semi-supervised tasks.

%


\bibliographystyle{icml2021}

\end{document}